# amLite: Amharic Transliteration Using Key Map Dictionary

Tadele Tedla
*St. Mary's University*
*Addis Ababa*
*Ethiopia*
ttedla@gmail.com

## Abstract

*amLite is a framework developed to map ASCII transliterated Amharic texts back to the original Amharic letter texts. The aim of such a framework is to make existing Amharic linguistic data consistent and interoperable among researchers. For achieving the objective, a key map dictionary is constructed using the possible ASCII combinations actively in use for transliterating Amharic letters; and a mapping of the combinations to the corresponding Amharic letters is done. The mapping is then used to replace the Amharic linguistic text back to form the original Amharic letters text. The framework indicated 97.7, 99.7 and 98.4 percentage accuracy on converting the three sample random test data. It is; however, possible to improve the accuracy of the framework by adding an exception to the implementation of the algorithm, or by preprocessing the input text prior to conversion. This paper outlined the rationales behind the need for developing the framework and the processes undertaken in the development.*

*Index Terms—Transliteration, Ethiopic, Localization, Amharic, CLIR, Query translation.*

## 1. Introduction

AMHARIC is a family of Semitic language widely spoken in Ethiopia. It is the second most spoken Semitic language in the world next to Arabic [1]. In addition, it is also serving as a national language in Ethiopia. As the result, materials produced in Amharic text using Amharic writing software are becoming abundant. The born Amharic digital texts include on-line newspapers, blogs, institutional or personal websites. The software helping the production of these text materials can be one of the available Amharic writing applications like Geez, Power Ge'ez, Visual Ge'ez, Agafari, Brana or other opensource packages.

To represent the phonemes of the language, Amharic uses its own set of alphabets known as 'fidel'. Including the vowels, there are a total of thirty five major letters each having up to seven minor variations formed by applying vowels on them [2]. The vowels are አ/a/ ኡ/u/ ኢ/i/ አ/A/ ኤ/E/ አ/e/ and ኦ/o/. For example በ/b/ which is often represented by the ASCII letter 'b' as shown in the bracket is considered as a major letter; and the variations formed by applying the vowels such as በ/be/ ቡ/bu/ ቢ /bi/ ባ/ba/ ቤ/bE/ and ቦ/bo/ are taken as minor letters in this paper. The representations shown within the brackets are one of the many ways of transliterating the alphabets in ASCII.

Transliteration of Amharic alphabets for computational analysis is mostly done using ASCII characters. Some of the reasons for this are, shipping of computers with ASCII keyboards, and computers are not yet capable of processing Amharic alphabets natively and uniformly. As a result, studies on Amharic language processing like [3] [4] [5] [6] are carried out using ASCII transliteratedAmharic. For example, [3] worked using the System for Ethiopic Representation in ASCII known by the abbreviation SERA transliteration scheme; whereas [4] tried to develop his own transliteration table which is none ASCII compliant. These development of personal transliteration scheme based on one's preference greatly challenges interoperability of the data among researchers of Amharic language. In addition, transliterating the data using ASCII every time a data is needed for computational analysis is a tedious, redundant task which creates a duplication of efforts. For example, [4] transliteration table maps the Amharic letter ዐ using /ä'/; that is, the letter ä plus a single quotation mark. The problem with this sort of representation is; letters like ä are found mostly on Nordic keyboards; which makes the process of using the transliterated texts made by his system difficult to process and implement on computers using standard keyboard software. In addition, ä is a Unicode character like the Amharic ዐ letter it represents. Hence, the benefit of representing a Unicode character with another Unicode character for linguistic analysis is a duplication of efforts. Furthermore, this paper is unable to find the transliteration scheme of [4] employed elsewhere for transliterating Amharic documents. These reasons make his transliteration table none useful for amLite key map dictionary construction.

On the other hand, the System for Ethiopic Representation in ASCII (SERA)[1] is a project initiated to standardize the transliteration of Amharic letters using ASCII. The consideration taken by the developers of SERA is, being able to type Amharic texts on a 101 keyboard which is presumed to be easy for computer translation. In addition, in SERA, most of the Amharic letters are represented using maximum of two ASCII characters. This makes SERA more transportable across computer systems. As a result, the data produced can easily be made interoperable among the Amharic language research community as well as computers processing the data. In addition, the popularity and availability of text transliterations using the scheme makes SERA part of amLite key map dictionary.

## 2. The Problem

The Amharic alphabets have no representation in the ASCII code table. As a result, the various Amharic writing software developers used their own ASCII letter combinations to represent the alphabets using ASCII keyboard. This often makes the different Amharic writing software to be exclusive and none interoperable. A text written using Visual Ge'ez; for example, may not be recognized by Power Ge'ez. The other problem is, most, if not all, of the Amharic language processing studies undertaken are done using transliteration. That is, an Amharic text is transliterated using for example, ASCII for linguistic processing or retrieval study. These transliterated versions of Amharic, however; have no standard [4] [6], and this makes the task of developing and sharing linguistic data very difficult. In addition, it is difficult to model the computational expenses of the different transliterations of a native Amharic text reliably. The problems posed by the inability to run standard measures due to the different transliteration scheme is a problem no one seems to consider the repercussions yet.

Table 1 shows the transliterations found in use to transliterate the Amharic word "ንጉሤ" in ASCII.

---

[1] http://www.abyssiniacybergateway.net/fidel/sera-94.html

Table 1
Possible ways of transliterating an Amharic word using ASCII

| ንጉሤ | ngu'sE (SERA form) |
| | ngussE |
| | ngu'see |
| | ngussee |
| | ngus2E |
| | ngus2ee |

One of the implications of having different transliteration scheme is the inability to have control over the internal representation of the letters on memory. That means, if transliteration is employed to process the example word given at table 1, the number of bytes required to represent the different transliterations of the same Amharic word varies. For example, the SERA implementation of the above name requires twenty seven bytes whereas the 'ngus2ee' requires twenty eight bytes of memory during computational operation like sorting on a 32 bit Intel desktop used for the experiment. Table 2 shows the number of bytes needed to represent the different transliteration of the Amharic word "ንጉሤ".

Table 2
Numbers of bytes need to represent the different transliterations of ንጉሤ

| Transliterations of ንጉሤ | Memory usage in Bytes (32 bit Intel desktop) |
|---|---|
| ngu'sE (SERA form) | 27 |
| ngussE | 27 |
| ngu'see | 28 |
| ngussee | 28 |
| Ngus2E | 27 |
| ngus2ee | 28 |

As shown in Table 2, some of the transliterations of the word needed more memory than others. The memory requirement grows linearly with the size of transliterated data. However; the Unicode implementation of the word in its native Amharic letters "ንጉሤ" for example, using a python utf-8 implementation; requires forty bytes of memory. An extra twelve bytes difference in a word accumulated can yield a significant performance difference on inputs of larger sizes. As a result, conducting a performance analysis on transliterated Amharic likely provides unreliable result. In addition, despite Unicode implementation variations, it is consistent to standardize and work on the word "ንጉሴ" in its native form than the six different variant transliterations.

The Amharic letters now have their own code points in Unicode, and they can be recognized by Unicode aware applications. Hence, there is no apparent benefit to be gained by working on transliterated Amharic linguistic data other than simplifying the characters to get the the support of of applications developed for ASCII. So, this paper tried to address these issues by developing an ASCII key map dictionary of the Amharic letters, so it can be used to convert the readily available linguistic data into Amharic letters automatically for reliable linguistic analysis and sharing of the data among researchers.

## 3. Related Work

The two major areas of research related to this study are Cross Language Information Retrial(CLIR) and query translation.

### A. Cross Language Information Retrieval

The Internet is enabling information to be available online in many languages. As a result, resources are becoming available in local languages of a nation other than English as once it is used to be. The availability of online resources in languages other than English for search initiates a need for CLIR [7]. The main goal of CLIR is to retrieve relevant documents irrespective of the language in whichdocuments are written [1] [8] [9]. For example, a search for the word "ግራዚኢኒ" is ideally expected to retrieve relevant documents to the term not only in Amharic but also in other languages. The transliteration of the above word "ግራዚኢኒ" using SERA, for instance, is "graziani" which is an exact English equivalence of the term in small caps. So, using transliteration, a search can retrieve the relevant documents from both Amharic and English language collections [10].

### B. Query Translation

A similar area of research to transliterating query for CLIR is query translation. Query translation relieves users from the burden of formulating queries in a non-native language for searching. Searching a monolingual English collection using Amharic queries can be possible with the help of query translation. A user query formulated in Amharic language, for example, can be translated to English before carrying out a search on English document collections. This can be done with the use of dictionaries [11] or other statistical methods [12]. Investigating which translation methods perform best for Amharic query translation is an interesting area of research requiring an independent study on itself. However; the transliteration dictionary developed in this study can be tailored to the use of query translation by mapping queries instead of letters to collections.

## 4. Operational Definition of Terms

**Transliteration:**
is a process of converting ASCII represented Amharictexts back to the original Amharic letters. For example converting 'ngussE' back to its Amharic letterrepresentation 'ንጉሴ' is transliteration.

**ASCII:**
an American Standard Code for Information Interchange, and it is an eight bit character representationscheme. It includes character representations from 0x00 to 0x7f; which are 128 characters in total.

**Code point:**
is an abstract way of assigning unique integer values to characters independent of their implementations. For example the Amharic letter 'ን' has a code point of u+1295.

**Malformed letter:**
an Amharic letter violating the correct spelling of a word. For example, the word ካረያ instead of ቃረያ is a malformed word. Both ካ and ቃ are valid Amharic letters but ካ in ካረያ is a malformed letter as it is not the correct Amharic word signifying an object in Amharic language. These are impossible to correct without employing a spell checker corpus. Hence, they cannot be considered as invalid output in this study. The other set of malformed letters comes from a valid input producing a wrong mapping and output. For example 'le2 se'at' should produce 'ለ 2 ሰዓት'; however, mapping of the longest combination 'e2' in this case is done first as it exists at the top of the mapping dictionary. As a result it produces 'ልዕ ሰዐት' which is not a valid Amharic word; and hence called malformed in this study.

**Unicode:**
is an international encoding standard representing Amharic letters within the range of u+1200 to u+137f code point. There are a total of three hundred eight four code points reserved for Amharic letters.

**Utf-8:**
a variable length encoding scheme which can encode all the three hundred eighty four Amharic Unicode code points using eight bit numbers.

# 5. amLite's Framework

This section describes the structure underlying amLite's development.

## A. The Key Map Dictionary

The first stage of amLite development is the construction of a key map dictionary. To construct the key map, the SERA mapping and other previous transliterated Amharic texts are referred. On the basis of which the ASCII transliteration keys are identified. The letters found in use are 'a', 'b', 'c', 'd', 'e', 'f', 'g', 'h', 'i', 'j', 'k', 'l', 'm', 'n', 'o', 'p', 'q', 'r', 's', 't', 'u', 'v', 'w', 'x', 'y', 'z', 'A', 'B', 'C', 'D', 'E', 'F', 'G', 'H', 'I', 'J', 'K', 'L', 'M', 'N', 'O', 'P', 'Q', 'R', 'S', 'T', 'U', 'V', 'W', 'X', 'Y', 'Z', ' ' ', '1', '2', '3', '4', '5', '6', '7', '8', '9', '0', '?', '!', '.', ':', ',', ';', '"', '|', '>', '<', '+', '-', and '*' ; which are seventy six characters in total. These seventy six characters are combined variably to represent the Amharic letters. For example, ኋ is found represented using 'hW', 'hWa', ''hW', ''hWa', 'hhW', 'hhWa', 'h2W', 'h2Wa'. Such varied options of transliterating an Amharic letter are identified, collected and included in the dictionary. The dictionary is a list containing the ASCII combinations and the corresponding Amharic letters. Table 3 shows a sample of the key map dictionary.

Table 3 Sample key map dictionary

| ASCII | Amharic |
|---|---|
| hW<br>hWa<br>'hW<br>'hWa<br>hhW<br>hhWa<br>h2W<br>h2Wa | ኋ |

Once the key map dictionary is constructed iteratively and incrementally, an algorithm for searching through the dictionary, and mapping of transliterated text into Amharic letter text is implemented

## B. The Algorithm

The algorithm follows a basic pattern of accepting one or more combinations of ASCII characters, process the characters and produce Amharic letters. In other words, read the letter 'b', for example, process it and produce the Amharic letter 'ብ'. However; the input in real cases, is not always as simple a letter as the example given above. The input can be a long sequence of transliterated random text spanning many pages. As a result, the algorithm should be able to find a way to tokenize the random input for automatic mapping. That means for example, given the input 'ngussE' there should be a way of automatically arranging it into 'n' 'gu' 'ssE' for correct conversion. The first stage of the algorithm development can simply be stated as follows.

**Input:** ASCII transliterated text (ngu'sE)
**Output:** Equivalent Amharic letter texts example (ንጉሴ)
**READ** (input, dictionary[ASCII_transliteration:Amharic_letter]);
**for** input in dictionary[ASCII_transliteration] **do**
    **REPLACE** input ← dictionary[Amharic_letter];
**end**
**WRITE** Amharic letters;

    **Algorithm 1**: amLite's basic algorithm

The READ operation reads the input into string and the dictionary; and passes them to the REPLACE operation which replaces the characters in the string with the corresponding Amharic letters and finally the WRITE operation writes the Amharic letters to the output file.

Since the dictionary is constructed over a long period of time, the arrangement of the ASCII representations and their respective Amharic letter is randomly spread inside the dictionary. That is, consider the letter ኋ and its transliterations. One of its transliteration might be located at the top where as the others at the middle or at the bottom of the dictionary. A simple mapping analysis without properly sorting the dictionary produced more incorrect outputs to the given sample strings. Sorting of the dictionary using the transliteration key length on the other hand corrected the outputs. As a result, sorting the dictionary using key length such that longer patters in the string will be automatically and randomly mapped has produced the most correct outputs. In addition, this implementationdecision controls the organization of the dictionary after the READ operation. Hence, sorting serves to token the input string automatically for mapping. The other problem is, the native Amharic letters are in Unicode, where as the transliterations exist in ASCII. It is necessary to label and normalize the ASCII columns of the dictionary as Unicode to make it usable for the subsequent operations like comparison, mapping and sorting. In addition, the input text is ASCII encoded or expected to be ASCII encoded. Explicitly declaring the encoding scheme of the the input text helps the algorithm to decode the text properly. Hence, the algorithm can handle the dictionary and input texts properly knowing their encoding. Then, it will be easy to replace the input text with Amharic letters found in the corresponding matching patterns of the encoded ASCII. For example, let's say we have an input transliterated Amharic text string ('ngussE') and thesorted dictionary will have entries ('ssE' → 'ሴ'; 'gu' → ' ጉ'; 'n' → 'ን') in order respectively for this particular instance.

So, the first REPLACE operation checks if the pattern 'ssE' is found in the input text. If it does, replaces 'ssE'

with 'ጒ' at its position and proceeds to see if the text has 'gu'. If it does, replaces it with 'ጉ' at position. In a similar way, the operation replaces 'n' with 'ን' and finally writes the Amharic letters with their respective position in the output file. To write the the output for viewing as native Amharic letter word 'ንጉሤ', the output needs to be explicitly encoded as utf-8.

The final algorithm is shown as Algorithm 2.
**Input**: ASCII transliterated text (ngu'sE)
**Output**: Equivalent Amharic letter texts example (ንጉሤ)
**READ** (input, dictionary[ASCII_transliteration:Amharic_letter]);
**NORMALIZE** dictionary[Unicode(ASCII_transliteration:Amharic_letter)];
**SORT** dictionary[ASCII_transliteration:Amharic_letter];
**DECODE** input;
**while** (!EOF) **do**
   **for** input in dictionary[ASCII_transliteration] **do**
   REPLACE
   input ← dictionary[Amharic_letter];
   **end**
**end**
**GENERATE** output ;
**ENCODE** output ;
**WRITE** output;
**Algorithm 2**: amLite's final algorithm

The READ operation reads the input text and the dictionary. The NORMALIZE dictionary operation makes sure that both the ASCII key and the value are in the proper encoding scheme. In our case, it is required to specify the Amharic letters and their corresponding ASCII transliterations as unicode. SORTing the dictionary is important to rearrange the random dictionary for searching and replacing automatically. The SORT brings longest keys to the top of the dictionary. In addition if two keys have the same length, then the one with smaller ASCII value will be sorted to the top of the dictionary. For example, if we consider the major Amharic letter 'ሀ', they might appear in the order of {'he': u'ሀ', 'hu': u'ሁ', 'hi':u'ሂ' , 'ha':u'ሃ', 'hE':u'ሄ', 'hee':u'ሄ', 'h':u'ህ', 'ho':u'ሆ'} in the dictionary. The SORT operation will arrange the letters as {'hee':u'ሄ', 'hE':u'ሄ', 'ha':u'ሃ', 'he': u'ሀ', 'hi':u'ሂ', 'ho':u'ሆ' , 'hu': u'ሁ' and 'h':u'ህ'}. Six of the transliterations have an equal transliteration key length but they are sorted in ascending order. The input text needs to be DECODEed as ASCII to get both the input and dictionary in their correct encoding for mapping and replacing. REPLACE will then start replacing the longest transliterating ASCII matching the input from the dictionary first and GENERATEs an equivalent Unicode representation for the input. To examine the REPLACE operation closely consider the input text 'ngussE'. The REPLACE will be done as (ngu ጉ), (n ጉ·ሤ) and (ንጉ·ሤ) respectively. ENCODE then encodes the output of GENERATE to the appropriate encoding supported in the system; utf-8 in our case so that it can be properly viewed after the WRITE operation. The algorithm is implemented using python programming language .

## 6. Test Data

The first set of test data is an ASCII transliterated Amharic words consisting of 32,482 words. Each line of the text contains a word. This is done to manually identify malformed words. The second test data is a transliterated poem written to commemorate the one hundredth historical victory over the invading Italy to Ethiopia on the battle of Adwa which contains about 276 lines of text. The third set of data is a recipe to make Injera -the common local food in Ethiopia- which has around 31 lines of transliterated Amharic text. The test data is organized into separate files as test1.txt, test2.txt and test3.txt and supplied as an input amLite.

Table 4 Summary of Experiment

| Input | Number of Words | Output |
|---|---|---|
| test1.txt | 32480 | test1out.txt |
| test2.txt | 1277 | test2out.txt |
| test3.txt | 123 | test3out.txt |

Table 5 Summary of malformed words in the output files

| Output file | Number of malformed or error output words |
|---|---|
| test1out.txt | 23 |
| test2out.txt | 5 |
| test3out.txt | 2 |

The count of malformed in test1out.txt is halted after checking the first one thousand words. This is because the cause of malformation kept repeating itself. That is, the algorithm replaces letters with longer ASCII key length first. As a result whenever there is a vowel following another vowel within a word such as 'meaza', like 'e' following 'a', 'ea' will be replaced first as 'ea' is found first than 'me' in the sorted key map dictionary. Hence 'ea' is replaced with 'ኧ' and 'm' with 'ም' and 'za' with 'ዛ' to become 'ምኧዛ' instead of the correct transliteration 'መአዛ'. A similar problem occurs with words like 'xi235', 'ke200xi' and 'yeInfalot'. The expected output was 'ሺ235', 'ከ200ሺ' and 'የእንፋሎት', however; the output produced ሽዒ35, 'ከዕ00ሺ' and 'ይዕንፋሎት' respectively which are allmalformed as the result of the reason explained above. The errors in test2out.txt and test3out.txt are also problems of similar causes.

Using the above table output, we have calculated the percentage of correct output produced by amLite. That is for test1.txt, test2.txt and test3.txt, the percentage of correct output is 97.7, 99.7 and 98.4 respectively. The percentage of accuracy can be improved processing the input prior to submission or adding an exception to the algorithm implementation.

## 7. Conclusion

The study achieved its objective of developing a scalable English ASCII to Amharic key map dictionary by con sulting transliterated texts, the Unicode standard, existing Amharic writing software key maps and the SERA. New key pairs can easily be added to the dictionary using (ASCII_key: Unicode_char). The algorithm is generic but needs an addition of exception to handle malformation during implementation. That can be done by controlling the input text having those repeated vowels as exceptions or by rearranging them down to their proper representation such as me-a-za instead of suppling the input as 'meaza' or else by analyzing the frequency of occurrence of the letters causing malformation in Amharic words to make an implementation decision.

Amharic language is one of the under resourced languages[2]. Finding linguistic corpus that can be used for various purposes of linguistic computation is a challenge. A method of integrating amLite with a spell checking dictionary[3] to match and suggest similar words is an interesting area of further study. Investigating an efficient method of correcting malformed words is another area of research which should further be investigated.

## Acknowledgments

The author would like to thank St.Mary's University for making the Internet available to its staff. Without which it would have been impossible to get the majority of the references used in this study. Though it is not his area of specialization, Assaye Legesse read the draft of this paper and gave stylistic comments. My wife Mekdes Bizuwork deserves to be acknowledged in many ways for the role she plays while undertaking projects since inception. The coffee she made was the strength it became for those after work hour endeavors of this project.

___

2 http://borel.slu.edu/crubadan/stadas.htm
3 http://www.cs.ru.nl/biniam/geez/

## References

[1] A. A. Argaw and L. Asker, "An amharic stemmer: Reducing words to their citation forms," *in Proceedings of the 2007 workshop on computational approaches to semitic languages*: Common issues and resources. Association for Computational Linguistics, 2007, pp. 104–110.

[2] B. Yemam., አጭርና ቀላል የአማርኛ ሰዋሰው. Addis Ababa, Ethiopia.:Alpha Printers, 2004.

[3] B. Gambäck, F. Olsson, A. A. Argaw, and L. Asker, "Methods for amharic part-of-speech tagging," *in Proceedings of the First Workshop on Language Technologies for African Languages.* Association for Computational Linguistics, 2009, pp. 104–111.

[4] M. Kassa Gobena, "Implementing an open source amharic resource grammar in gf," 2011.

[5] A. SOLOMON, "Unsupervised machine learning approach for word sense disambiguation to amharic words," 2011.

[6] S. Teklu, Automatic Categorization Of Amharic News Text: A Machine Learning Approach.LAP Lambert Academic Publishing, 2012.

[7] L. Ballesteros and W. B. Croft, "Phrasal translation and query expansion techniques for cross-language information retrieval," in ACM SIGIR Forum, vol. 31, no. SI. ACM, 1997, pp. 84–91.

[8] D. W. Oard, D. He, and J. Wang, "User-assisted query translation for interactive cross-language information retrieval," Information Processing & Management, vol. 44, no. 1, pp. 181–211, 2008.

[9] J. G. Carbonell, Y. Yang, R. E. Frederking, R. D. Brown,Y. Geng, and D. Lee, "Translingual information retrieval: A comparative evaluation," 1997.

[10] A. A. Argaw and L. Asker, "Amharic-english information re trieval," in Evaluation of Multilingual and Multi-modal Information Retrieval. Springer, 2007, pp. 43–50.

[11] A. A. Argaw, L. Asker, R. Cöster, and J. Karlgren, "Dictionarybased amharic–english information retrieval," in Multilingual Information Access for Text, Speech and Images. Springer, 2005, pp. 143–149.

[12] J. Gao, J.-Y. Nie, E. Xun, J. Zhang, M. Zhou, and C. Huang, "Improving query translation for cross-language information retrieval using statistical models," in Proceedings of the 24th annual international ACM SIGIR conference on Research and development in information retrieval. ACM, 2001, pp. 96–104.